\documentclass{article}

\usepackage{microtype}
\usepackage{graphicx}
\usepackage{subfigure}
\usepackage{booktabs} 

\usepackage{hyperref}
\usepackage{nicefrac}

\usepackage[accepted]{icml2024}

\usepackage{amsmath}
\usepackage{amssymb}
\usepackage{mathtools}
\usepackage{amsthm}

\usepackage[capitalize,noabbrev]{cleveref}

\theoremstyle{plain}

\theoremstyle{definition}

\theoremstyle{remark}

\usepackage[textsize=tiny]{todonotes}

\usepackage{notation}

\newtheorem*{proposition*}{Proposition}

\icmltitlerunning{Improving GFlowNets with Monte Carlo Tree Search}

\begin{document}

\twocolumn[
\icmltitle{Improving GFlowNets with Monte Carlo Tree Search}




\begin{icmlauthorlist}
\icmlauthor{Nikita Morozov}{hse,sk}
\icmlauthor{Daniil Tiapkin}{paris,CNRSSACLAY}
\icmlauthor{Sergey Samsonov}{hse}
\icmlauthor{Alexey Naumov}{hse,ras}
\icmlauthor{Dmitry Vetrov}{cub}

\end{icmlauthorlist}

\icmlaffiliation{hse}{HSE University, Moscow, Russia}
\icmlaffiliation{sk}{Skoltech, Moscow, Russia}
\icmlaffiliation{paris}{CMAP – CNRS – {\'E}cole polytechnique – Institut Polytechnique de
Paris, 91128, Palaiseau, France}
\icmlaffiliation{CNRSSACLAY}{Université Paris-Saclay, CNRS, LMO, 91405, Orsay, France}
\icmlaffiliation{cub}{Constructor University, Bremen}
\icmlaffiliation{ras}{Steklov Mathematical Institute of Russian Academy of Sciences}

\icmlcorrespondingauthor{Nikita Morozov}{nvmorozov@hse.ru}

\icmlkeywords{Machine Learning, ICML}

\vskip 0.3in
]



\printAffiliationsAndNotice{}  

\begin{abstract}
Generative Flow Networks (GFlowNets) treat sampling from distributions over compositional discrete spaces as a sequential decision-making problem, training a stochastic policy to construct objects step by step. Recent studies have revealed strong connections between GFlowNets and entropy-regularized reinforcement learning. Building on these insights, we propose to enhance planning capabilities of GFlowNets by applying Monte Carlo Tree Search (MCTS). Specifically, we show how the MENTS algorithm~\cite{xiao2019maximum} can be adapted for GFlowNets and used during both training and inference. Our experiments demonstrate that this approach improves the sample efficiency of GFlowNet training and the generation fidelity of pre-trained GFlowNet models.
\end{abstract}


\vspace{-0.4cm}

\section{Introduction}
\label{sec:intro}\vspace{-0.05cm}

Generative Flow Networks (GFlowNets, \citealp{bengio2021flow}) are models designed to sample compositional discrete objects, such as graphs, from distributions defined by unnormalized probability mass functions. They achieve this by training a stochastic policy to generate objects through a sequence of constructive actions to match the desired distribution. GFlowNets have been successfully applied in various areas, including biological sequence design \cite{jain2022biological}, large language model (LLM) fine-tuning \citep{hu2023amortizing}, combinatorial optimization \cite{zhang2023solving}, neural architecture search \citep{chen2023order}, and causal discovery \cite{atanackovic2024dyngfn}.

GFlowNets incorporate many concepts and techniques from reinforcement learning (RL). Recent works \cite{tiapkin2024generative, mohammadpour2024maximum, deleu2024discrete} have shown that the GFlowNet learning problem can be reformulated as an RL problem with entropy regularization \cite{neu2017unified, geist2019theory}. These findings opened a direct way to apply many existing RL algorithms \cite{schulman2017equivalence, haarnoja2017reinforcement, haarnoja2018soft} to GFlowNets, and our work follows this path.

Monte Carlo Tree Search (MCTS) is a well-known method for solving planning problems \cite{coulom2006efficient, kocsis2006bandit}. Prominent examples of RL algorithms utilizing MCTS include \texttt{AlphaGo} \citep{silver2016mastering} and \texttt{AlphaZero} \citep{silver2017general}, which combine MCTS with deep neural networks to achieve superhuman performance in games like Go, chess, and Shogi. MCTS algorithms typically require knowledge of the environment's underlying dynamics or can be paired with a neural network-based simulator, as seen in \texttt{MuZero}-type approaches \cite{schrittwieser2020mastering}, resulting in a complicated algorithm. Fortunately, GFlowNets fall into the first category because the directed acyclic graph (DAG) environments they operate in are integral to the algorithm's design and, moreover, deterministic. Thus, \textit{the ability to simulate any trajectory in a deterministic DAG environment makes the idea of enhancing the planning abilities of GFlowNets with MCTS very natural.}

We focus on the Maximum Entropy for Tree Search (\texttt{MENTS}, \citealp{xiao2019maximum}) algorithm, an MCTS algorithm that estimates entropy-regularized Q-values. This entropy-regularized nature of this algorithm allows it to be directly applied to the framework of GFlowNets. \textbf{We outline our contributions as follows:} \emph{i)} We show how \texttt{MENTS} coupled with \texttt{SoftDQN}~\cite{haarnoja2017reinforcement} can be applied to GFlowNets at both training and inference stages, \emph{ii)} we experimentally demonstrate how improved planning capabilities can benefit GFlowNets.

\vspace{-0.15cm}
\section{Background}
\label{sec:background}

\subsection{GFlowNets}
Suppose we have a finite space \(\mathcal{X}\) and a black-box non-negative function \(R \colon \mathcal{X} \to \mathbb{R}_{\geq 0}\), which we will call the \textit{GFlowNet reward}. Our goal is to sample objects from \(\mathcal{X}\) with probabilities \(R(x) / \mathrm{Z}\), where \(\mathrm{Z} = \sum_{x \in \mathcal{X}} R(x)\) is an unknown normalizing constant.

Consider a finite directed acyclic graph (DAG) \(\mathcal{G} = (\mathcal{S}, \mathcal{E})\), where \(\mathcal{S}\) is a state space and \(\mathcal{E} \subseteq \mathcal{S} \times \mathcal{S}\) is a set of edges. Non-terminal states correspond to "incomplete" objects, with an empty object denoted as \(s_0\), and edges represent adding new components to these objects. Every state can be reached from \(s_0\), which has no incoming edges. Terminal states are "complete" objects and coincide with \(\mathcal{X}\). Let \(\mathcal{T}\) denote the set of all complete trajectories \(\tau = \left(s_0 \to s_1 \to \ldots \to s_{n_{\tau}}\right)\) in the graph, where \(\tau\) is a sequence of transitions \(s_i \to s_{i + 1} \in \mathcal{E}\) from \(s_0\) to some terminal state \(s_{n_{\tau}} \in \mathcal{X}\).

Next, we introduce probability distributions over the children of each state \(P_F(s_t \mid s_{t-1})\) and the parents of each state \(P_B(s_{t-1} \mid s_t)\), called the \textit{forward policy} and the \textit{backward policy}, respectively. The main goal is to find a pair of policies such that the induced distributions over complete trajectories in the forward and backward directions coincide:
\begin{equation}\label{eq:tb}
     \prod_{t=1}^{n_\tau} P_F\left(s_{t} \mid s_{t-1}\right) = \frac{\cR(s_{n_\tau})}{\rmZ} \prod_{t=1}^{n_\tau} P_B\left(s_{t-1} \mid s_{t}\right)\, \; \forall \tau \in \cT.
\end{equation}
This is known as the \textit{trajectory balance constraint} \cite{malkin2022trajectory}. If this constraint is satisfied for all complete trajectories, sampling a trajectory in the forward direction using \(P_F\) will result in a terminal state being sampled with probability \(\cR(x)/\mathrm{Z}\).

In practice, GFlowNet is a model that parameterizes the forward policy (and possibly other components) trained to minimize an objective function that enforces the constraint \eqref{eq:tb} or an equivalent one. Among existing training objectives, \textit{Subtrajectory Balance} (\texttt{SubTB}, \citealp{madan2023learning}) has been shown experimentally to have superior performance across various tasks. Notably, the backward policy can either be trained alongside the forward policy or fixed, for example, to be uniform over the parents of each state. For any fixed backward policy, there exists a unique forward policy that satisfies \eqref{eq:tb} \cite{malkin2022trajectory}. For further details on GFlowNets, we refer to \cite{bengio2023gflownet}.

\subsection{GFlowNets as Soft RL}
\label{sec:gfn_as_rl}

In contrast to the classical RL formulation of reward maximization, entropy-regularized RL (\citealt{neu2017unified, geist2019theory,haarnoja2017reinforcement}, also known as soft RL) augments the value function by Shannon entropy $\cH$: 
\begin{equation}\label{eq:regularized_value_def}
    \textstyle V^\pi_{\lambda}(s) \triangleq \E_{\pi}\bigg[ \sum\limits_{t=0}^\infty \gamma^t (r(s_t,a_t) + \lambda \cH(\pi(s_t))) | s_0 = s\bigg],
\end{equation}
where $\lambda$ is a regularization coefficient. Similarly, we can define regularized Q-values  $Q^\pi_{\lambda}(s,a)$ as an expected (discounted) sum of rewards augmented by Shannon entropy given a fixed initial state $s_0=s$ and action $a_0 = a$. A regularized optimal  policy $\pistar_{\lambda}$ is a policy that maximizes $V^{\pi}_{\lambda}(s)$ for any initial state $s$.

Let $\Vstar_{\lambda}$ and $\Qstar_{\lambda}$ be the value and the Q-value of the optimal policy $\pistar_{\lambda}$ correspondingly. Then Theorem~1 and 2 by \cite{haarnoja2017reinforcement} imply the following system relations for any state-action pair $s,a$ for a deterministic environment
\begin{align}\label{eq:soft_bellman_equations}
    \begin{split}
        \Qstar_{\lambda}(s,a) &= r(s,a) + \gamma \logsumexp_{\lambda}\left( \Qstar_{\lambda}(s', \cdot)\right)\,,
    \end{split}
\end{align}
where $s'$ is a next state after taking an action $a$ in a state $s$, $\logsumexp_\lambda(Q(s', \cdot)) \triangleq \lambda \log \big( \sum_{a'} \exp\left\{ Q(s', a')/\lambda \right\}\big)$. Then the optimal policy can be computed as $\pistar_\lambda(\cdot \mid s) \triangleq \softmax(\Qstar_{\lambda}(s,\cdot) / \lambda).$ 

\cite{tiapkin2024generative} showed that the problem of GFlowNet forward policy training given a fixed backward policy can be equivalently formulated as an entropy-regularized RL problem. This reduction involves adding an absorbing state \(s_f\) to the GFlowNet DAG, with edges from terminal states to \(s_f\) and a loop $s_f \to s_f$. A deterministic Markov Decision Process (MDP) is then constructed from the DAG, where states correspond to DAG states and actions correspond to edges (or, equivalently, to next possible states). RL rewards are set for all edges as follows:
\begin{equation}\label{eq:def_mdp_reward}
        r(s,s') \triangleq \begin{cases}
            \log P_B(s \mid s') & s \not \in \cX \cup \{s_f\}, \\
            \log \cR(s) & s \in \cX, \\
            0 & s = s_f\,.
        \end{cases}
\end{equation}
Theorem 1 of \cite{tiapkin2024generative} states that the optimal policy \(\pi^*_1(s' \mid s)\) in this MDP, with  \(\lambda\) set to 1 and \(\gamma = 1\), coincides with the GFlowNet forward policy \(P_F\) (which is uniquely defined by \(P_B\) and \(R\)).

This reduction enables the direct application of soft RL algorithms to GFlowNet training. \cite{tiapkin2024generative} applied the classical \texttt{SoftDQN} algorithm \cite{haarnoja2017reinforcement} and demonstrated its efficiency. Essentially, a neural network is trained to predict optimal regularized Q-values for all transitions using the following objective:
\begin{equation}\label{eq:softdqn_loss}
    \big( Q_\theta(s, s') - \log P_B(s \mid s') - \logsumexp\left(Q_{\bar{\theta}}(s', \cdot)\right) \big)^2,
\end{equation}
where $\logsumexp(Q_{\bar{\theta}}(s', \cdot)) \triangleq \logsumexp_1(Q_{\bar{\theta}}(s', \cdot))$ 
 is replaced with $\log \cR(s')$ if $s' \in \cX$, and $\bar\theta$ are parameters of a target network that is updated with weights $\theta$ from time to time. The corresponding policy is computed as $\pi_\theta(\cdot \mid s) = \softmax(Q_{{\theta}}(s, \cdot))$. The model can be either trained on-policy by optimizing the loss over complete trajectories sampled from $\pi_{\theta}$ or utilize a replay buffer.

\section{Method}
\label{sec:method}

In RL planning, the agent needs to determine the optimal action to maximize future rewards in a large state space. The simplest method is to train a Q-network to predict the expected future rewards for each action and choose the one with the highest predicted Q-value. However, this approach depends heavily on the Q-network's approximation capabilities and may not fully leverage the problem's structure. In contrast, MCTS algorithms look multiple steps ahead to evaluate the future state of the environment better. MCTS incrementally builds a look-ahead tree, balancing the exploration-exploitation trade-off during navigation in the tree~\cite{kocsis2006bandit}. Each new node added to the tree is evaluated either through a Monte Carlo simulation or neural network prediction, and this information is backpropagated along the path to the root.

In GFlowNets, the planning problem differs because we need to determine not just a single optimal action but the optimal distribution over possible actions (forward policy). This can be achieved by training a Q-network, as the optimal policy is the softmax of optimal entropy-regularized Q-values. However, using look-ahead information from MCTS can provide better estimates of Q-values in a similar fashion to the RL setting. Therefore, we propose a direct adaptation of the \texttt{MENTS} algorithm \cite{xiao2019maximum}, which aims to improve the estimation of optimal entropy-regularized Q-values. Below, we describe how \texttt{MENTS} can be applied to GFlowNet inference and training on top of \texttt{SoftDQN}, following the paradigm described in Section~\ref{sec:gfn_as_rl}.

\subsection{MENTS for GFlowNets}

\textbf{Inference stage.} Suppose we have a pre-trained neural network $Q_\theta$ that predicts soft Q-values. The root of the look-ahead tree corresponds to the current DAG state $s_{\mathrm{root}}$. For each node of the tree, we store a visit count $N(s)$, and for each edge $s \to s'$, we store an estimate $Q_{\mathrm{tree}}(s, s')$ of the regularized Q-value. 

During each round of MCTS, we sample a path from the root to some leaf of the tree by sequentially sampling a child from the tree policy, that is, a softmax policy with respect to $Q_{\mathrm{tree}}$ with an additional $\varepsilon$-greedy exploration:
\begin{equation}\label{eq:tree_policy}
    \pi_{\mathrm{tree}}(\cdot \mid s) = (1 - p_s) \softmax(Q_{\mathrm{tree}}(s, \cdot)) +  p_s 
 \cdot \cU(C(s))\,,
\end{equation}
where $\cU(C(s))$ is a uniform distribution over the children of $s$ denoted by $C(s)$, and $p_s = \varepsilon |C(s)| / \log(N(s) + 2)$, where $\varepsilon$ is an exploration hyperparameter. 

Let $(s_{1}, s_2, \ldots, s_T)$ be the sampled path, where $s_1 = s_{\mathrm{root}}$ and $s_T$ is a leaf. Then, we add new nodes and edges to the tree corresponding to the children of $s_T$ in the GFlowNet DAG $\cG$. For each added child $s' \in C(s_T)$, we initialize $N(s') = 0$ and $Q_{\mathrm{tree}}(s_T, s') = Q_{\theta}(s_T, s')$. Then, for each node in the path, we update $N(s_i) = N(s_i) + 1$, and for each edge in the path from last to first, we update the Q-value estimate, following the optimality condition~\eqref{eq:soft_bellman_equations} 
\begin{equation}\label{eq:ments_update}
Q_{\mathrm{tree}}(s_i, s_{i + 1}) = \log P_B(s_i \mid s_{i + 1}) + \logsumexp\left(Q_{\mathrm{tree}}(s_{i + 1}, \cdot)\right).
\end{equation}
A special case arises when $s_T$ is a terminal state of $\cG$. No nodes will be added to the tree in this case, and as for updating $Q_{\mathrm{tree}}(s_{T-1}, s_{T})$, there are two options. The first option is to replace $\logsumexp\left(Q_{\mathrm{tree}}(s_{T}, \cdot)\right)$ with $\log \cR(s_T)$ (since $\log \cR(s_T)$ coincides with $\Vstar_1(s_T)$, see~\citealp{tiapkin2024generative}). However, in potential scenarios with no access to GFlowNet reward $\cR$ during inference or its calculation being very expensive (e.g., drug discovery, see~\citealp{jain2023gflownets}), this option may not be very practical. The second option is to skip the update of $Q_{\mathrm{tree}}(s_{T-1}, s_{T})$, leaving it as it was initialized by $Q_{\theta}(s_{T-1}, s_{T})$. Then the algorithm only requires access to $Q_{\theta}$, $\log P_B$, and the structure of $\cG$, making it applicable in all practical cases. All experiments in Section~\ref{sec:experiments} are carried out with this option.

After all rounds of MCTS, we have an estimate $Q_{\mathrm{tree}}(s_{\mathrm{root}}, s')$ for each child $s'$ of the root, and the resulting forward policy can be obtained as $\softmax(Q_{\mathrm{tree}}(s_{\mathrm{root}}, \cdot))$. The next state is sampled from this policy, and the tree's root is changed to the corresponding child, possibly already having a non-empty subtree. Note that the number of times $Q_{\theta}$ is evaluated is upper bounded by the number of MCTS rounds (assuming $Q_{\theta}$ takes a state as an input and outputs predictions for all possible actions). In practice, we fix the maximum visit count of the root $N(s_{\mathrm{root}})$ as a hyperparameter; thus, the number of rounds can vary depending on the number of visits to the state before it becomes the root. 

An important point is that the presented algorithm does not require $\cG$ itself to be a tree and can work in arbitrary GFlowNet environments. If a state in $\cG$ is reached by a number of different paths during MCTS, there will be multiple nodes in the tree corresponding to the same state. Appendix~\ref{app:algo} presents a detailed pseudocode of the algorithm and its connection to GFlowNet flow functions.

\textbf{Training stage.} Consider \texttt{SoftDQN} training objective \eqref{eq:softdqn_loss}. It can be viewed as fitting $Q_{\theta}(s, s')$ on a one-step \texttt{MENTS} estimate calculated using the current target network $Q_{\bar{\theta}}$. However, one can run multiple rounds of MCTS to obtain better targets for fitting the Q-network, which allows us to utilize MCTS for training $Q_{\theta}$. The training objective becomes 
\begin{equation}\label{eq:softdqn_mcts_loss}
    \left ( Q_\theta(s, s') - Q_{\mathrm{tree}}(s, s')\right)^2,
\end{equation}
where $Q_{\mathrm{tree}}(s, s')$ is obtained by applying \texttt{MENTS} with the current target network $Q_{\bar{\theta}}$ instead of a fixed pre-trained one. Since we do not provide access to GFlowNet rewards during MCTS, an exception is a loss for transitions into terminal states $s' \in \X$, which we take to be  
\begin{equation}\label{eq:softdqn_mcts_loss_terminal}
    \left ( Q_\theta(s, s') - \log P_B(s \mid s') - \log \cR(s') \right)^2.
\end{equation}

Such a choice also allows for a more straightforward comparison with other methods in terms of the number of calls to $\cR(x)$  made during training.


\section{Experiments}
\label{sec:experiments}

We carry out experimental evaluation on hypegrid~\cite{bengio2021flow} and bit sequence~\cite{malkin2022trajectory} environments following similar experimental setups to~\cite{tiapkin2024generative}. Along with \texttt{SoftDQN} and \texttt{MENTS}, we provide \texttt{SubTB}~\cite{madan2023learning} as a baseline. In all experiments, $P_B$ is fixed to be uniform. Appendix~\ref{app:exp} contains additional experimental details and runtime measurements for the compared algorithms.

\begin{figure}[!t]
    \centering
    \includegraphics[width=0.4951\linewidth]{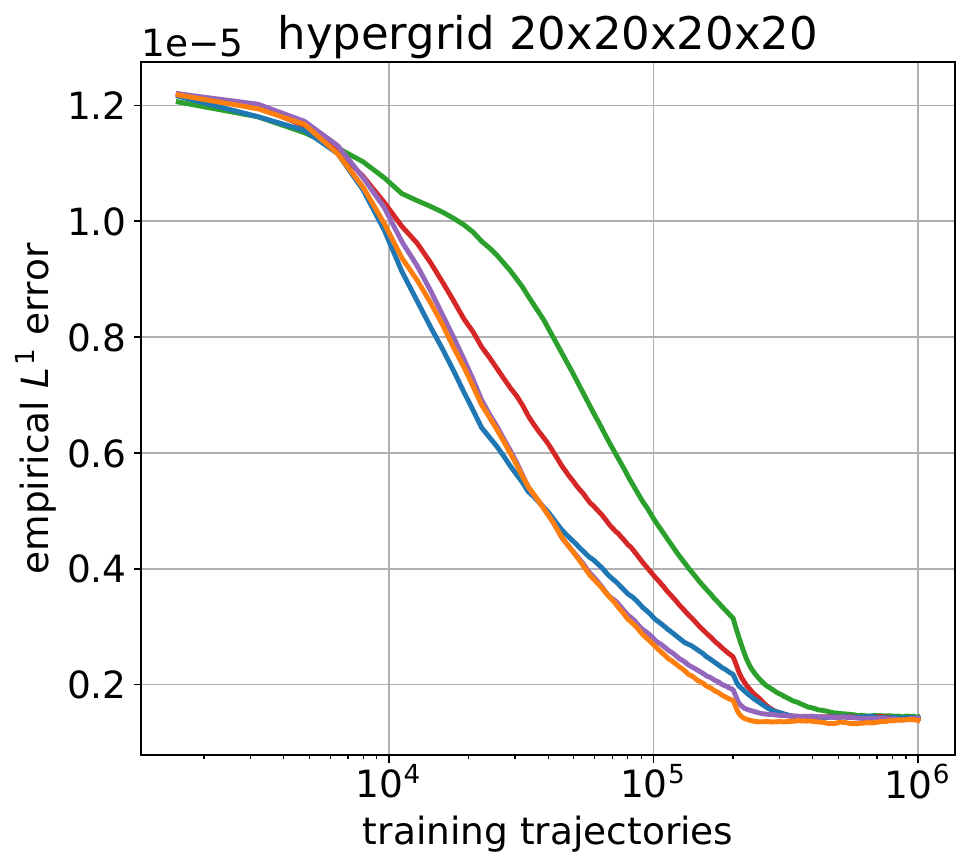}
    \includegraphics[width=0.4951\linewidth]{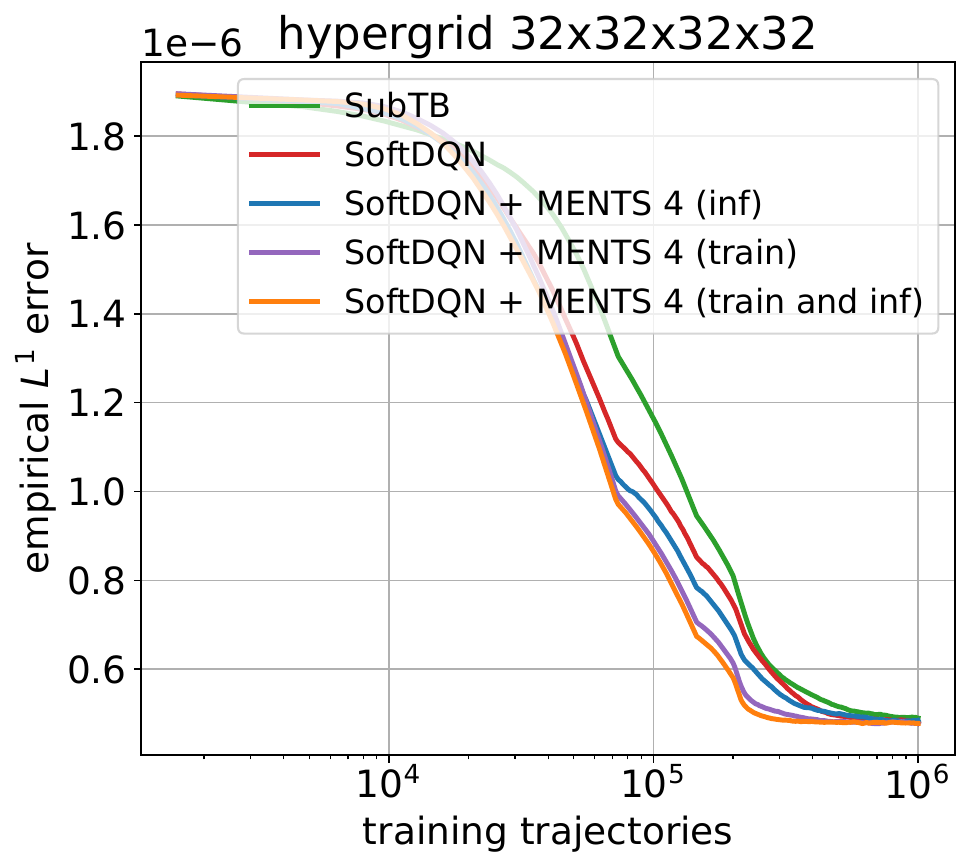}
    \includegraphics[width=0.4951\linewidth]{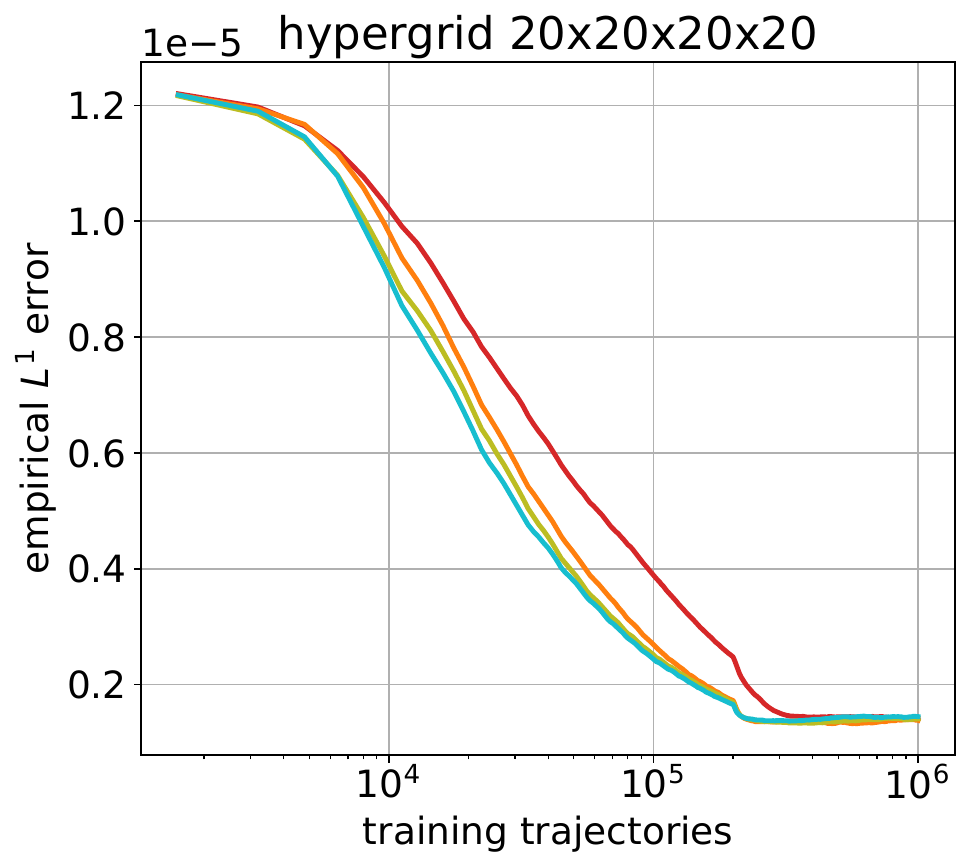}
    \includegraphics[width=0.4951\linewidth]{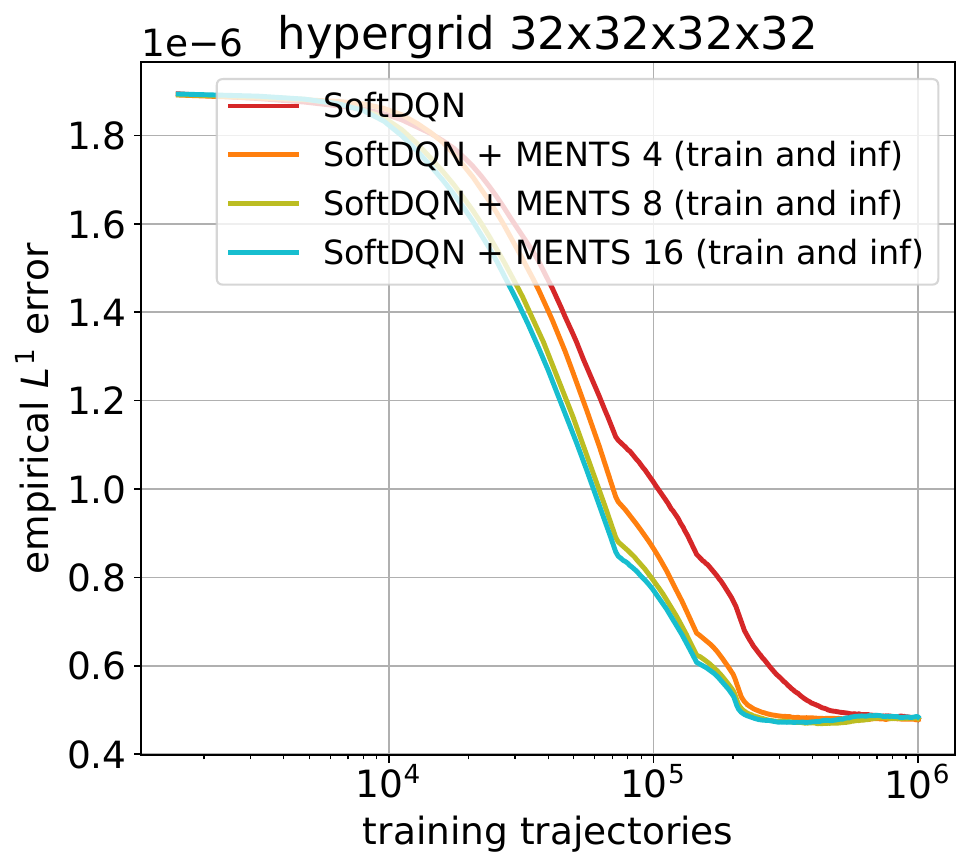}
    
    \caption{$L^1$ distance between target and empirical sample distributions over the course of training on the hypergrid environment. Numbers next to \texttt{MENTS} in the legend correspond to maximum number of MCTS rounds $N(s_{\mathrm{root}})$.} 
    \label{fig:hypergrid}
    \vspace{-0.3cm}
\end{figure}

\subsection{Hypergrid Environment}

The set of states corresponds to two copies of points (non-terminal and terminal) with integer coordinates inside a 4-dimensional hypercube with side length $H$. The allowed actions are to increase on coordinate by $1$ without exiting the grid and to move to a terminal copy of the state. Initial state $s_0$ is $(0, 0, 0, 0)$. The reward has modes near the corners of the grid, separated by wide troughs with a very small reward. All models are
parameterized by MLP with one-hot encoded inputs. 

We study 3 setups: 1) a model is trained with vanilla \texttt{SoftDQN} and evaluated with \texttt{MENTS}; 2) a model is trained with \texttt{MENTS} targets, but the trained policy is evaluated without MCTS; 3) \texttt{MENTS} is applied for both training and evaluation. In contrast to~\cite{tiapkin2024generative}, we do not use replay buffers for training, instead optimizing the loss across trajectories sampled from the current model. As a metric we use $L^1$ distance between the true reward distribution $\cR(x)/\rmZ$ ($\rmZ$ can be computed exactly
since environments are small) and an empirical distribution
of $2 \cdot 10^5$ GFlowNet samples.

Figure~\ref{fig:hypergrid} presents the results. We can see that in all setups \texttt{MENTS} offers a stable improvement to the speed of convergence in comparison to vanilla \texttt{SoftDQN} in terms of number of sampled trajectories, which coincides with the number of calls to $R(x)$. The best results are obtained when \texttt{MENTS} is applied for both training and inference of the model. Remarkably, using \texttt{MENTS} to compute targets for the training of $Q_{\theta}$ provides a noticeable boost even when the model is evaluated without MCTS (setup number 2). Increasing the number of MCTS rounds is also beneficial.

\begin{figure}[!t]
    \centering 
    \includegraphics[width=0.4951\linewidth]{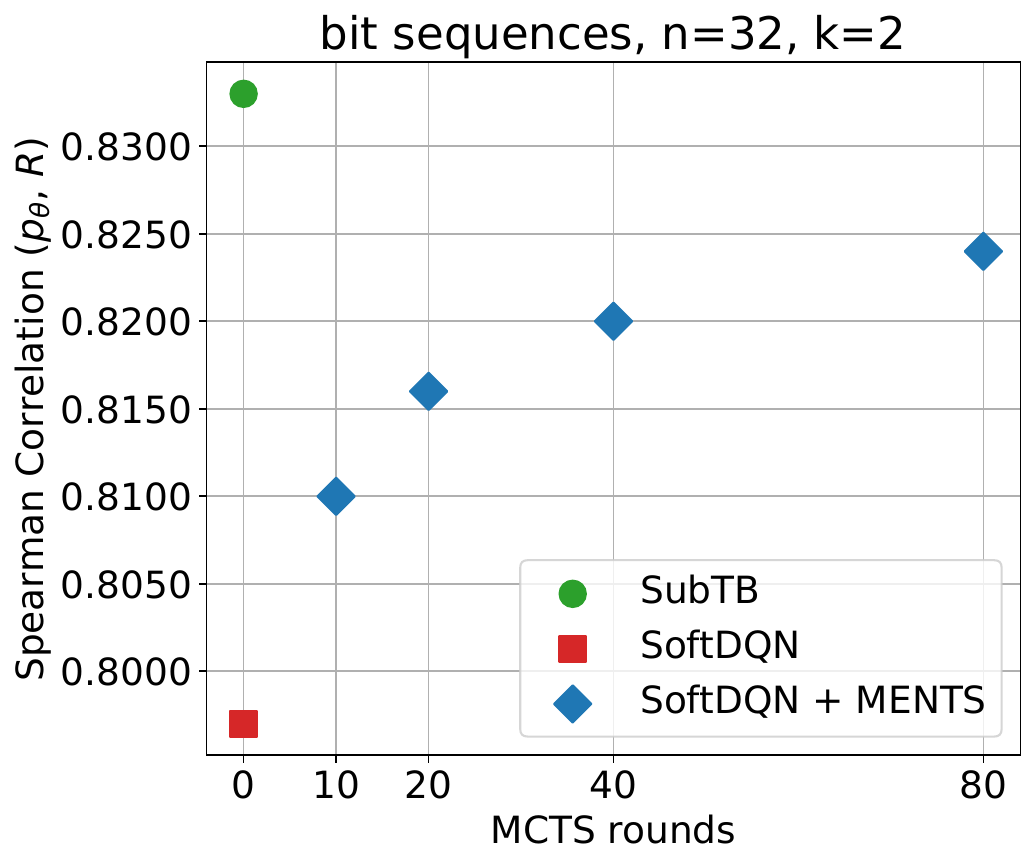}
    \includegraphics[width=0.4951\linewidth]{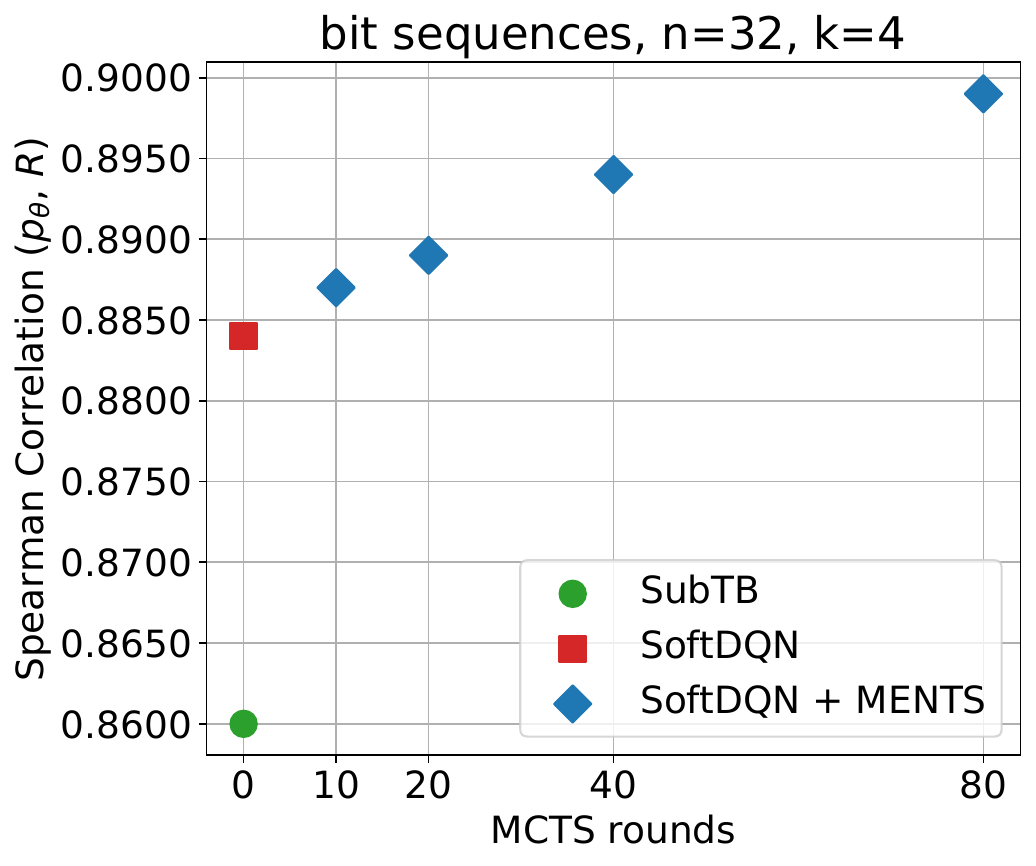}
    \includegraphics[width=0.4951\linewidth]{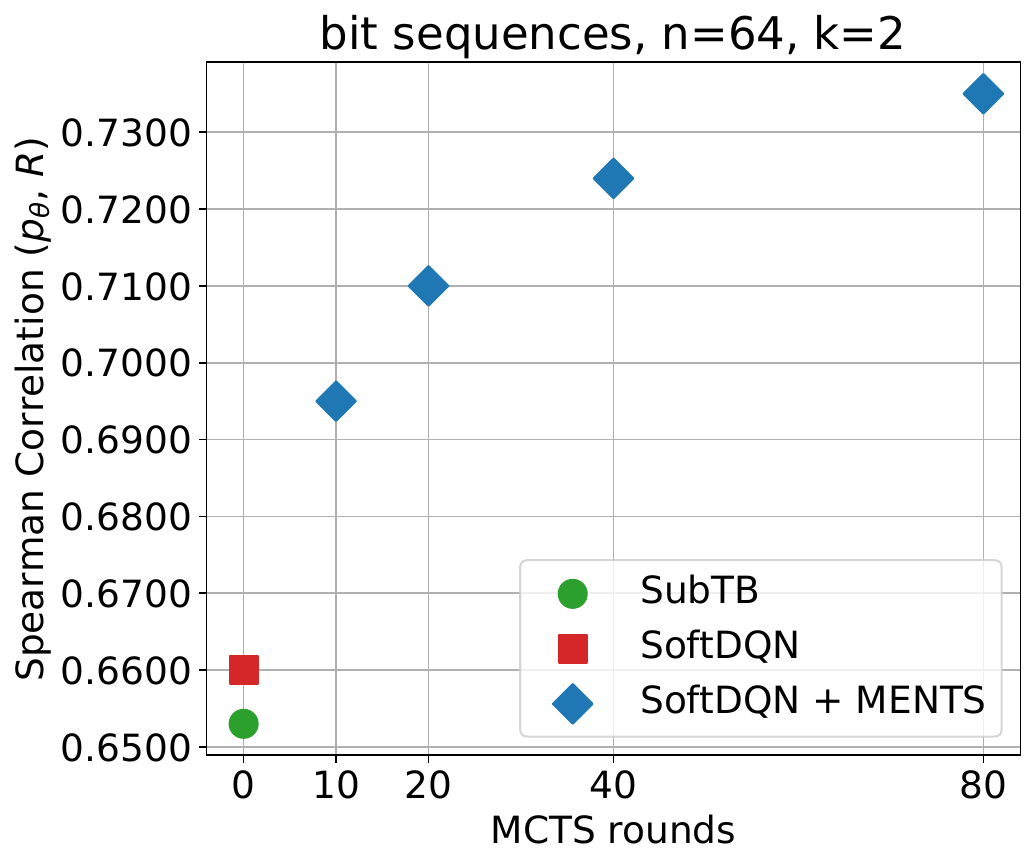}
    \includegraphics[width=0.4951\linewidth]{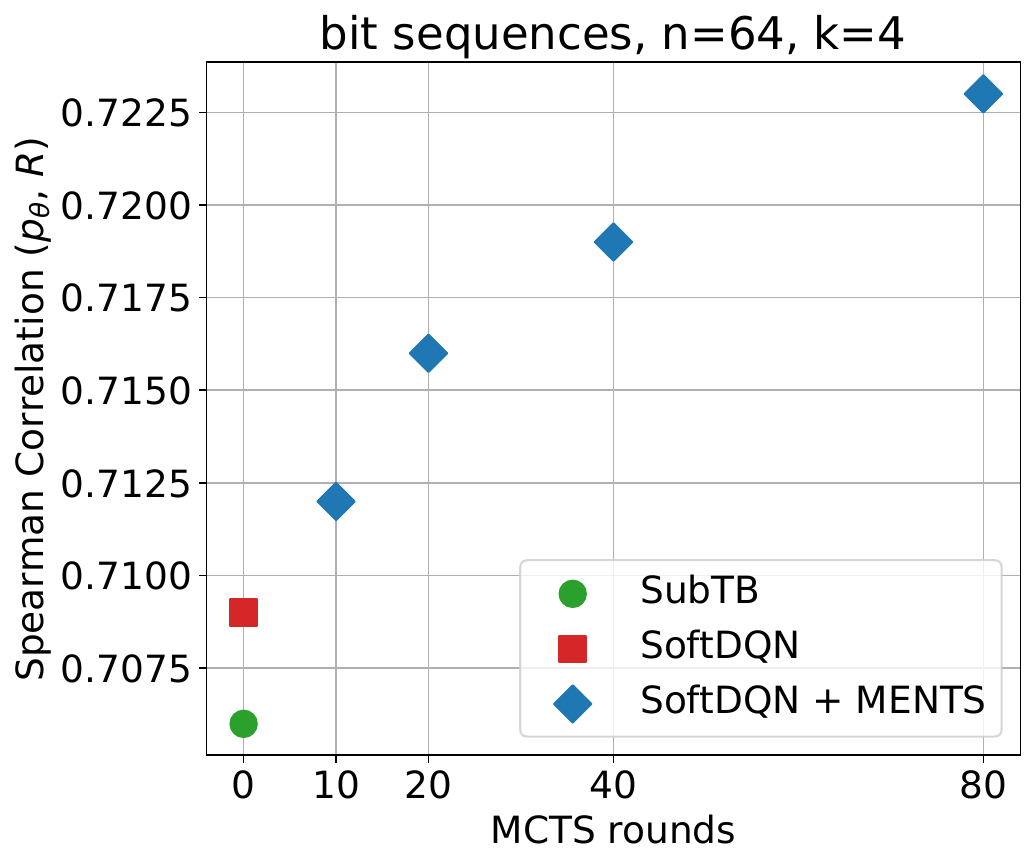}
    
    \caption{Spearman correlation between $R$ and $P_{\theta}$ on a test set for varying $n$ and $k$ in the bit sequence generation task. \texttt{MENTS} is used here only at the inference stage.}
    \label{fig:bits}
    \vspace{-0.3cm}
\end{figure}

\subsection{Bit Sequence Generation}
\label{sec:exp}

The goal is to generate binary strings of some fixed length $n$. Hyperparameter $k \mid n$ is introduced, and the string is split into $n/k$ segments of length $k$. Each state corresponds to a sequence of $n/k$ words; each word is either an empty word $\oslash$ or one of $2^k$ possible $k$-bit words. $s_0$ corresponds to a sequence of empty words. Possible actions are to take any position with an empty word and replace it with any $k$-bit word. Terminal states contain no empty words and coincide with binary strings of length $n$. ${\cR(x) = \exp(-2 \cdot \min_{x' \in M} d(x, x'))}$, where $M$ is a set of modes and $d$ is Hamming distance. We use this environment to examine the performance of MCTS in a more challenging setup with larger state and action spaces (up to $\approx 2 \cdot 10^{22}$ states and $256$ actions in our experiments). 

Here we train $Q_{\theta}$ with \texttt{SoftDQN} paremeterized by Transformer~\citep{vaswani2017attention} and utilize \texttt{MENTS} \textit{only during inference}. Following~\cite{tiapkin2024generative} we compute Spearman correlation on a test set of strings between $R$ and an estimate of sampling probability $P_{\theta}$. The results are presented in Figure~\ref{fig:bits}. In all configurations, enhancing \texttt{SoftDQN} with \texttt{MENTS} improves the reward correlation in comparison to vanilla \texttt{SoftDQN}, although the improvement is relatively small in some cases. It also outperforms \texttt{SubTB} in 3 out of 4 cases. The metric steadily rises with the increase of the number of MCTS rounds. 

\section{Conclusion}
\label{sec:conclusion}


In this paper, we proposed to apply \texttt{MENTS} \cite{xiao2019maximum} algorithm with \texttt{SoftDQN} \cite{haarnoja2017reinforcement} to GFlowNet training and inference. Our experimental results demonstrated the benefits of incorporating MCTS planning for amortized sampling, suggesting new research directions. Future work could explore other MCTS-type approaches, validate them in other domains, and apply MCTS on top of other GFlowNet algorithms, e.g. \texttt{SubTB}~\cite{madan2023learning}.

\section*{Acknowledgements}

The work of Nikita Morozov, Sergey Samsonov and Alexey Naumov was supported by the grant for research centers in the field of AI provided by the Analytical Center for the Government of the Russian Federation (ACRF) in accordance with the agreement on the provision of subsidies (identifier of the agreement 000000D730321P5Q0002) and the agreement with HSE University No. 70-2021-00139. The work of Daniil Tiapkin was supported by the Paris Île-de-France Région in the framework of DIM AI4IDF. This research was supported in part through computational resources of HPC facilities at HSE University~\citep{kostenetskiy2021hpc}.


\bibliography{main}
\bibliographystyle{icml2024}

\newpage
\appendix
\onecolumn





\section{Algorithm Details}\label{app:algo}

Algorithm~\ref{gfn_ments} presents a detailed pseudo-code for sampling a trajectory with \texttt{MENTS} applied on top of a GFlowNet pre-trained with \texttt{SoftDQN}.

\begin{algorithm}[H]
   \caption{Inference of SoftDQN + MENTS}
   \label{alg:example}
\begin{algorithmic}[1]
   \STATE {\bfseries Input:} SoftDQN pre-trained $Q_{\theta}$, maximum number of MCTS rounds $N_{\mathrm{max}}$, exploration parameter $\varepsilon$, GFlowNet backward policy $P_B$
   \STATE Initialize $s_{\mathrm{root}} = s_0$, $N(s_{\mathrm{root}}) = 0$
   \REPEAT
   \WHILE{$N(s_{\mathrm{root}}) < N_{\mathrm{max}}$}
   \STATE Initialize $\mathrm{path} = \{ s_{\mathrm{root}} \}$
   \WHILE{$\mathrm{path}.\mathrm{last}$ is not a leaf}
   \STATE Initialize $s = \mathrm{path}.\mathrm{last}$
   \STATE Compute $p_s = \varepsilon |C(s)| / \log(N(s) + 2)$
   \STATE Compute $\pi_{\mathrm{tree}}(\cdot \mid s) = (1 - p_s) \cdot \softmax(Q_{\mathrm{tree}}(s, \cdot)) +  p_s\cdot \cU(C(s))$
   \STATE Sample $s' \sim \pi_{\mathrm{tree}}(\cdot \mid s)$
   \STATE Append $s'$ to $\mathrm{path}$
   \STATE Update $N(s') = N(s') + 1$
   \ENDWHILE
   \STATE Initialize $s_{\mathrm{leaf}} = \mathrm{path}.\mathrm{last}$
    \IF{$s_{\mathrm{leaf}} \not\in \cX$}
   \FORALL{$s' \in C(s_{\mathrm{leaf}})$}
   \STATE Add $s'$ to the tree 
   \STATE Initialize $N(s') = 0$
   \STATE Initialize $Q_{\mathrm{tree}}(s_{\mathrm{leaf}}, s') = Q_{\theta}(s_{\mathrm{leaf}}, s')$
   \ENDFOR
   \ENDIF
   \FOR{$i = \mathrm{path}.\mathrm{size} - 1$ {\bfseries to} $1$}
   \IF{$\mathrm{path}_{i + 1} \not\in \cX$}
   \STATE Update $Q_{\mathrm{tree}}(\mathrm{path}_i, \mathrm{path}_{i + 1}) = \log P_B(\mathrm{path}_i \mid \mathrm{path}_{i + 1}) + \logsumexp\left(Q_{\mathrm{tree}}(\mathrm{path}_{i + 1}, \cdot)\right)$
   \ENDIF
   \ENDFOR
   \ENDWHILE
   \STATE Compute $P_F(\cdot \mid s_{\mathrm{root}}) = \softmax(Q_{\mathrm{tree}}(s_{\mathrm{root}}, \cdot))$
   \STATE Sample $s_{\mathrm{next}} \sim P_F(\cdot \mid s_{\mathrm{root}})$
   \STATE Delete everything from the tree except the subtree of $s_{\mathrm{next}}$
   \STATE Update $s_{\mathrm{root}} = s_{\mathrm{next}}$
   \UNTIL{$s_{\mathrm{root}}$ corresponds to a terminal state $x \in \cX$}
   \STATE {\bfseries Output} terminal state $x$
\end{algorithmic}
\label{gfn_ments}
\end{algorithm}

\vspace{-0.3cm}
\subsection{Connection to GFlowNet State and Edge Flows}

Suppose we have forward and backward policies that satisfy trajectory balance constraints~\eqref{eq:tb}. Then, we have a fixed distribution over complete trajectories 
\begin{equation}
    P(\tau) =  \prod_{t=1}^{n_\tau} P_F\left(s_{t} \mid s_{t-1}\right) = \frac{\cR(s_{n_\tau})}{\rmZ} \prod_{t=1}^{n_\tau} P_B\left(s_{t-1} \mid s_{t}\right).
\end{equation}

GFlowNet literature often operates with \textit{flows} functions~\cite{bengio2023gflownet}. \textit{Markovian flow} in this case is a function $F\colon \mathcal{T} \to \mathbb{R}_{\geq 0}$ that coincides with unnormalized probability of sampling a trajectory $F(\tau) = \rmZ \cdot P(\tau)$. Since for any fixed $P_B$ and $R$ there exists a unique $P_F$ satisfying~\eqref{eq:tb}, any fixed $P_B$ and $R$ also define a unique Markovian flow. \textit{State flows} and \textit{edge flows} are defined as $F(s) = \sum_{\tau \ni s} F(\tau)$, $F\left(s \to s^{\prime}\right) = \sum_{\tau \ni (s \to s') } F(\tau)$ correspondingly, and coincide with unnormalized probabilities that a trajectory passes through some state/edge.

\textit{Flow matching constraint} states that for any state that is not $s_0$ or terminal
\begin{equation}
    F(s) = \sum_{s \to s'} F(s \to s') = \sum_{s'' \to s} F(s'' \to s),
\end{equation}
while for $s_0$ and terminal states, only one of the two equalities holds.

$P_F$ and $P_B$ can be computed in terms of state and edge flows
\begin{equation}\label{eq:pfpb}
P_F\left(s^{\prime} \mid s\right)=\frac{F\left(s \rightarrow s^{\prime}\right)}{F(s)}, \quad P_B\left(s \mid s^{\prime}\right)=\frac{F\left(s \rightarrow s^{\prime}\right)}{F\left(s^{\prime}\right)}.
\end{equation}

\vspace{-0.2cm}
Let us go back to the RL interpretation. In addition to the optimal policy, Theorem 1 of \cite{tiapkin2024generative} connects state and edge flows with optimal entropy-regularized values and Q-values, stating
\begin{equation}
    \Vstar_1(s) = \log F(s), \quad \Qstar_1(s, s') = \log F(s \to s').
\end{equation}
In this interpretation, \eqref{eq:soft_bellman_equations} transforms into 
\begin{align}\label{eq:gfn_bellman}
\begin{split}
     \log F(s \to s') &= \log P_B(s \mid s') + \logsumexp(\log F(s' \to \cdot)) \\ &= \log P_B(s \mid s') + \log \sum_{s' \to s''} F(s' \to s'') \\ &= \log P_B(s \mid s') + \log F(s'),
\end{split}
\end{align}
which can also be obtained from~\eqref{eq:pfpb}. The equation on the optimal policy $\pistar_1(\cdot \mid s) = \softmax(\Qstar_1(\cdot \mid s))$ transforms into
\begin{equation}
    P_F(s' \mid s) = \exp\left(\log F(s \to s') - \logsumexp(\log F(s \to \cdot))\right) = \exp\left(\log F(s \to s') - \log F(s)\right),
\end{equation}
which also coincides with the equation on $P_F$ from~\eqref{eq:pfpb}.

If we try to look into the algorithm described in Section~\ref{sec:method} in terms of flow functions, one can interpret that it applies MCTS on top of a neural network $\log F_{\theta}(s \to s')$ and tries to estimate $\log F_{\mathrm{tree}}(s \to s')$ for edges in the tree. The update formula~\eqref{eq:ments_update} actually coincides with~\eqref{eq:gfn_bellman}: 
\vspace{-0.1cm}
\begin{equation}
    \log F_{\mathrm{tree}}(s \to s') = \log P_B(s \mid s') + \logsumexp(\log F_{\mathrm{tree}}(s' \to \cdot)).
\end{equation}

\vspace{-0.3cm}
\section{Experimental Details}\label{app:exp}


We utilize PyTorch~\citep{paszke2019pytorch}, and our implementations are based upon the published code of~\cite{tiapkin2024generative}. We implement \texttt{MENTS} in C++ for better performance.

\vspace{-0.1cm}
\subsection{Hypergrid}\label{app:exp_grid}

The reward at a terminal state $s$ with coordinates $(s^1, \ldots, s^D)$ is defined as
\begin{equation*}
\cR(s) = 10^{-3} + 0.5 \cdot \prod_{i = 1}^D \mathbb{I}\left[0.25 < \left|\frac{s^i}{H-1}-0.5\right|\right] + 2 \cdot \prod_{i = 1}^D \mathbb{I}\left[0.3 < \left|\frac{s^i}{H-1}-0.5\right| < 0.4\right].
\end{equation*}

\vspace{-0.2cm}
We use similar hyperparameters to previous works~\cite{bengio2021flow, malkin2022trajectory, madan2023learning, tiapkin2024generative}. All models are parameterized by MLP with 2 hidden layers and 256 hidden units. We use Adam optimizer with a learning rate of $10^{-3}$ and a batch size of 16 trajectories. We take \texttt{SubTB} hyperparameter $\lambda = 0.9$. The difference from~\cite{tiapkin2024generative} is that for \texttt{SoftDQN}, we do not use a replay buffer and use MSE loss instead of Huber. We use hard updates for the target network~\citep{mnih2015human} with a frequency of $3$ iterations. For \texttt{MENTS} we take $\varepsilon=0.01$. We perform hypergrid experiments on CPUs. 

In Table~\ref{tab:speed}, we measure the runtime of the algorithms during training and inference. As expected, the speed of \texttt{MENTS} decreases with the increase of the number of rounds due to additional $Q_{\theta}$ evaluations. However, we note that training with \texttt{MENTS} (4 rounds) runs faster than with \texttt{SubTB} and has better convergence than both \texttt{SubTB} and vanilla \texttt{SoftDQN} (see Figure~\ref{fig:hypergrid}).

\begin{table}[H]
\small
\centering
\caption{Training and inference speed on hypergrid environment measured on Apple M1 CPU. One iteration coincides with a batch of $16$ trajectories in all cases. The lower training speed of \texttt{SubTB} in comparison to \texttt{SoftDQN} is due to the fact that the number of terms in its loss is quadratic in the trajectory length.} \small
\vskip 0.12in
\begin{tabular}{l|cc}
Method                  & Training & Inference \\
\hline
SubTB  & 8.5 it/s                 & 35.6 it/s             \\
SoftDQN & 20.5 it/s                & 35.8 it/s             \\
SoftDQN + MENTS 4 & 12.3 it/s                 & 14.0 it/s             \\
SoftDQN + MENTS 8   & 8.1 it/s                & 9.2 it/s              \\
SoftDQN + MENTS 16   & 5.3 it/s                & 6.3 it/s             
\end{tabular}
\label{tab:speed}
\end{table}

\subsection{Bit Sequences}\label{app:exp_bit}

The set of modes $M$ is constructed as defined in~\cite{malkin2022trajectory}, and we use the same size $|M| = 60$. Take $H=\left\{' 00000000^{\prime}, ' 11111111^{\prime}, ' 11110000^{\prime}, 00001111^{\prime}, ' 00111100^{\prime}\right\}$. Then, each sequence in $M$ is constructed by randomly taking $n/8$ elements from $H$ with replacement and concatenating them. The test set for computing reward correlations is constructed by taking a mode and flipping $i$ random bits in it, which is done for each mode and each $0 \le i < n$.

We use the same Monte Carlo estimate for $P_{\theta}$ as in~\cite{tiapkin2024generative}:
\begin{align*}
    P_{\theta}(x) = \mathbb{E}_{P_B(\tau \mid x)} \frac{P_F(\tau \mid \theta)}{P_B(\tau \mid x)} \approx \frac{1}{N} \sum\limits_{i = 1}^N \frac{P_F(\tau^i \mid \theta)}{P_B(\tau^i \mid x)}, \;\; \tau^i \sim P_B(\tau \mid x).
\end{align*}

All models are parameterized by Transformer~\cite{vaswani2017attention} with 2 hidden layers, 8 attention heads and 64 hidden dimension. We use Adam optimizer with a learning rate of $10^{-3}$ and a batch size of $16$. For \texttt{SubTB} we tune $\lambda$ from $\{0.9, 1.1, 1.9\}$. For training $\texttt{SoftDQN}$, we use hard updates for the target network with a frequency of $5$ iterations and use Huber loss following~\cite{tiapkin2024generative}. We also utilize a prioritized replay buffer~\cite{schaul2016prioritized} with the same hyperparameters as in~\cite{tiapkin2024generative}. For \texttt{MENTS} we take $\varepsilon=0.001$. We use NVIDIA A100 GPUs for bit sequence experiments.

In this case the runtime cost of tree manipulation in MCTS is dominated by the cost of $Q_{\theta}$ forward passes, thus the inference speed decreases proportionally to the number of MCTS rounds.

\end{document}